\documentclass[letterpaper, 10 pt, journal, twoside]{IEEEtran}

\IEEEoverridecommandlockouts % for using \thanks
\usepackage[pagebackref=true,colorlinks,citecolor=brown]{hyperref}
\usepackage{booktabs}
\usepackage{amsmath}
\usepackage{amssymb}
\usepackage{macros}
\usepackage{graphicx}
\usepackage{svg}
\usepackage{subcaption}
\usepackage{makecell}

\usepackage{enumitem}
\usepackage[font=small]{caption}
\usepackage{float}
\usepackage{dblfloatfix}
\usepackage{adjustbox}
\usepackage{pdfpages}
\usepackage{placeins}

\usepackage{tikz}
\usetikzlibrary{positioning, arrows.meta, shapes.geometric}
\usepackage{wrapfig}
\usepackage{xspace}
\usepackage{todonotes}

\renewcommand{\paragraph}[1]{\vspace{.1em}\noindent\textbf{#1}.}

\usepackage{algorithm}
\usepackage{algorithmic}

\makeatletter
\newcommand\fs@betterruled{%
  \def\@fs@cfont{\bfseries}\let\@fs@capt\floatc@ruled
  \def\@fs@pre{\vspace*{7pt}\hrule height.8pt depth0pt \kern2pt}%
  \def\@fs@post{\kern2pt\hrule\relax}%
  \def\@fs@mid{\kern2pt\hrule\kern2pt}%
  \let\@fs@iftopcapt\iftrue}
\floatstyle{betterruled}
\restylefloat{algorithm}
\makeatother

\usepackage{pgfplots}
\usepackage{xcolor}
\pgfplotsset{compat=newest}

\newcommand{\taskimg}[2]{%
\begin{tabular}{@{}c@{}}
\includegraphics[width=0.072\textwidth]{#1}\\
{\scriptsize #2}
\end{tabular}
}

\begin{document}

% paper title
\title{\LARGE \bf Adaptive Mean Flow for Responsive Closed-Loop Robot Control}

\author{Aksel Vaaler$^{1}$, Marco Job$^{1}$, Christian Holden$^{1}$, and Olav Egeland$^{1}$
    \thanks{$^{1}$ Department of Mechanical and Industrial Engineering,
            Norwegian University of Science and Technology (NTNU)
            {\tt\small \{aksel.vaaler, marco.job, christian.holden, olav.egeland\}@ntnu.no}}%
    \thanks{Code: \href{https://github.com/akselva/Adaptive-mean-flow-RoboticIL}{github.com/akselva/Adaptive-mean-flow-RoboticIL}}
}

\maketitle

\begin{abstract}
Diffusion- and flow-based robot policies have recently become widespread in robotic Imitation Learning (IL) due to their high performance and ability to model continuous and multimodal distributions. However, the iterative denoising procedure used by these models introduces significant prediction latency, hindering high-frequency closed-loop robot control and leading to jittery, unstable motion when frequent updates to the robot's action predictions are used. Therefore, it is common practice to train models to predict chunks of actions that can be executed sequentially without feedback, even when this reduces responsiveness and may mean the most recent state information is not used. In this article, we present Adaptive Mean Flow (AMF), a flow-based IL method that enables smooth and responsive, fully closed-loop robot control. AMF uses Mean Flow, which is an accelerated form of Flow Matching (FM), to minimize prediction latency. To ensure smoothness and consistency across predictions, AMF uses a corrupted version of the trajectory from the previous step when predicting new robot actions, with the signal-to-noise ratio increasing over the time parameter of the trajectory. This discourages large changes in the prediction from one step to the next, while allowing freedom to adapt the predictions for future steps. We evaluate AMF across a wide range of simulated and real robot tasks and demonstrate significantly improved performance compared with baselines. Code: \href{https://github.com/akselva/Adaptive-mean-flow-RoboticIL}{github.com/akselva/Adaptive-mean-flow-RoboticIL}.
\end{abstract}

\IEEEpeerreviewmaketitle
\vspace{3pt}

%----INTRODUCTION---%
Diffusion models \cite{ho2020denoising, pmlr-v37-sohl-dickstein15} and Flow Matching (FM) models \cite{lipman2023flow} have recently become widespread in robotic Imitation Learning (IL) \cite{diffusionsurveywolf2025}. These models capture complex, high-dimensional distributions and are well-suited to model diverse, human-collected robot demonstration trajectories. Diffusion- and flow-based methods work by gradually transforming (typically Gaussian) noise into approximate samples from the data distribution in a process referred to as denoising \cite{ho2020denoising}. This iterative procedure yields high sample quality at the cost of slower inference speed. To take full advantage of the capacity afforded by these models while minimizing the impact of slow inference speed, it is proposed in \cite{chi2023diffusion} to predict sequences of future robot actions rather than just the action at the current timestep. This allows for faster execution of robot actions than the prediction speed of the policy, since action predictions from the previous control step can be sequentially passed to the robot while the policy calculates the next action sequence. This technique is called \textit{action chunking} \cite{zhao2023learning}, where a ``chunk'' refers to the actions in a predicted sequence that the robot executes before replanning. 

Another benefit of action chunking is improved temporal consistency during deployment \cite{chi2023diffusion}. While the demonstration data may exhibit diverse behavior and multimodality on a high level, each individual trajectory sample that the model is trained on will typically be smooth and self-consistent. This means that subsequent robot actions within a single predicted action sequence are likely to be consistent, while actions from different predicted sequences might diverge. For this reason, it is not uncommon to use a longer action chunk length than what is strictly required, given the inference latency of the diffusion/flow model. On the other hand, using longer chunk lengths decreases the policy's responsiveness and means that actions are not always taken based on the most recent information, which can be detrimental, especially in the face of environmental disturbances \cite{liu2025bidirectional}. This can be viewed as a fundamental trade-off between reactivity and consistency. Bridging this gap in diffusion and flow-based robotic IL requires policies that 1) minimize prediction latency to allow faster feedback and 2) maintain smooth and consistent behavior between control steps, while remaining responsive to changes in the environment.

\begin{figure}[t]
    \centering
    \includegraphics[width=.36\textwidth]{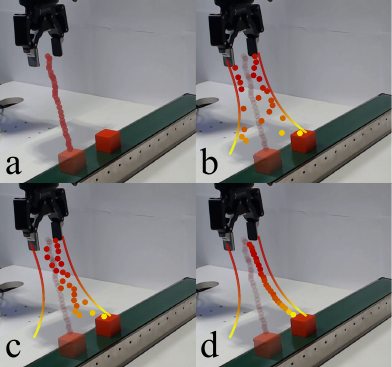}
    \caption{Trajectory generation with Adaptive Mean Flow visualized for a real-world conveyor belt picking task. \textbf{a-b)} Noisy points are sampled in the vicinity of the trajectory computed at the previous step. The noise increases for points along the path, corresponding to predictions farther into the future. \textbf{c-d)} The AMF denoising network computes a new trajectory by modifying the noisy sequence.}
    \label{fig:amf_rollout_fig}
    \vspace{-10pt}
\end{figure}

To address these challenges, we propose \textit{Adaptive Mean Flow} (AMF). AMF utilizes Mean Flow (MF) \cite{geng2025meanflowsonestepgenerative}, which allows sampling high-quality action sequences from noise in a single denoising step, reducing inference latency. This is crucial for allowing rapid feedback control and high responsiveness. To maintain consistency with previous robot actions while remaining adaptive to the environment's state, we draw inspiration from \cite{høeg2024streamingdiffusionpolicyfast,chenforcing2024} and let the noise level vary dynamically across robot actions in a sequence. During inference, instead of predicting new action sequences from pure noise, AMF reuses the previously predicted sequence as a starting point and corrupts this sequence with progressively increasing noise. This encourages near-future action predictions to remain close to previous predictions, while allowing actions farther in the future more freedom to adjust in response to the environmental state, as visualized in Fig.~\ref{fig:amf_rollout_fig}. We evaluate AMF on a comprehensive set of simulated and real-world robotic manipulation tasks, and show that it exhibits smooth and responsive robot control, outperforming the baselines in both static and dynamic environments.

The contributions of this paper are:
\begin{itemize}
    \item We present Adaptive Mean Flow (AMF), a novel robotic Imitation Learning policy that enables smooth, responsive closed-loop robot control.
    \item We empirically validate the proposed method on a diverse set of simulated and real-world robot tasks that require precision, consistency, and responsiveness to be solved reliably.
    \item We investigate key algorithmic and architectural design choices to optimize AMF performance.
\end{itemize}

\section{Related Work}

 \subsection{Speeding up diffusion and flow inference}
 Diffusion and flow models rely on iterative refinement (denoising) to produce high-quality output samples, which means that multiple forward passes of the network are needed per model prediction. Considerable research has been conducted to accelerate the inference process while maintaining quality. For applications with real-time demands, such as robotics, the deterministic Denoising Diffusion Implicit Models (DDIM) sampler proposed in \cite{song2021denoising} has largely been preferred in favor of the stochastic Denoising Diffusion Probabilistic Models (DDPM) sampler originally proposed in \cite{ho2020denoising}, as it generally requires fewer denoising steps to produce good samples. Consistency Policy \cite{prasadconsistency} can produce high-quality robot actions with very few denoising steps by utilizing the diffusion distillation framework developed in \cite{pmlr-v202-song23a}. Similarly, One-Step Diffusion Policy \cite{wang2025onestep} proposes a distillation scheme that enables training a student model in a single denoising step to mimic the action predictions of a pre-trained Diffusion Policy \cite{chi2023diffusion} network. Streaming Diffusion Policy \cite{høeg2024streamingdiffusionpolicyfast} and Diffusion Forcing \cite{chenforcing2024} train the networks using varying noise levels across the action sequence. This allows inference using a ``stair-case'' noise structure, where action predictions for near-future time steps are initialized with less noise and thus require fewer denoising steps.
 
 A flow model can itself be viewed as a faster alternative to diffusion, typically requiring far fewer integration steps compared with diffusion and DDPM sampling \cite{lipman2023flow}. However, induced curvature in the probability flow field used to generate samples means that several ODE integration steps are still required \cite{geng2025meanflowsonestepgenerative}. Analogous to Consistency Policy \cite{prasadconsistency}, Consistency Flow Policy \cite{zhangconsistencyflowpolicy2025} trains the network with additional consistency constraints on the flow ODE \cite{yang2024consistencyfm} to enable action generation with fewer integration steps. Streaming Flow Policy (SFP) \cite{jiang2025streaming} takes a different approach by formulating action trajectories as flows. Starting from the last robot action, SFP treats a step along the flow ODE as a step along the forward robot trajectory, meaning the next robot action is ``ready'' after a single integration step.

Our AMF method utilizes Mean Flow \cite{geng2025meanflowsonestepgenerative}, which was also used in \cite{Shengmeanflowtames2026}, and enables single-step action sequence prediction of high quality without the need for distillation, additional consistency losses, or complex inference algorithms. 

 \subsection{Consistency-reactivity trade-off in robotic Imitation Learning}
 The aforementioned works all address the problem of speeding up diffusion/flow inference, which offers immediate benefits in terms of smoother policy execution \cite{jiang2025streaming, høeg2024streamingdiffusionpolicyfast}. However, they still use action chunking to ensure consistent robot motion and thus do not fully utilize the responsiveness enabled by fast inference. Comparatively few works directly address the consistency-reactivity trade-off between action chunking and fully closed-loop, single-step action execution. However, there is a growing interest, particularly in developing test-time algorithms to improve the performance of pre-trained chunking-based policies during deployment. Bidirectional Decoding \cite{liu2025bidirectional} and Past-Token Prediction \cite{torne2025ptp} introduce algorithms that select the next action sequence from a set of candidates, with selection criteria that aim to balance consistency with the previous prediction and responsiveness to the most recent observation. Self-Guided Action Diffusion \cite{malhotra2025selfguidedactiondiffusion} and Real-Time Chunking \cite{black2025realtimeexecutionactionchunking} use weighted distances from the previous prediction to the current prediction to steer a diffusion/flow model during the denoising process, producing action sequences that more closely align with the trajectory prediction from the previous step. Temporal Action Selection \cite{weng2025temporalactionselectionaction} also employs a pre-trained IL policy, but uses online RL fine-tuning to train an action selector network to adaptively update the action trajectory based on current and previous predictions from the pre-trained policy.

 In contrast, our AMF method uses no fine-tuning or additional test-time compute. By training the model on action sequences with dynamic per-step signal-to-noise ratio, flexibility is retained with respect to controlling the behavior of the policy. During deployment, AMF predicts actions rapidly in closed-loop, maintaining consistency simply by initializing the noisy samples around the previous action trajectory.

 \section{Background}

 \textbf{Flow Matching:} Flow Matching \cite{lipman2023flow, liu2023flow} is a framework for learning velocity fields that transport samples from a known (typically Gaussian) prior distribution $p_{\text{prior}}$ into approximate samples from an unknown data distribution $p_{\text{data}}$. Given $x \sim p_{\text{data}}$ and $e \sim p_{\text{prior}}$, a linear transport schedule is defined as $z_s = sx + (1-s)e$ for $s \in [0,1]$. Taking the derivative with respect to $s$ yields the \textit{conditional velocity} $v_c = x - e$. Flow Matching involves regressing a neural network $v_\theta$ towards the conditional velocity:
 \begin{equation}\label{eq:cfm_loss}
     \mathcal{L} = \mathbb{E}_{s,x,e} \| v_\theta (z_s,s) - (x - e) \|^2.
 \end{equation}

It can be shown that the minimization of Eq.~\ref{eq:cfm_loss} is equivalent to the approximation of the \textit{marginal velocity} $v(z_s,s) = \mathbb{E} [v_c | z_s]$ \cite{lipman2023flow}. With a trained network $v_\theta(z_s,s)$, synthetic samples $z_1 \sim q \approx p_{\text{data}}$ can be generated by integrating the ODE $\frac{d}{ds}z_s = v_\theta(z_s,s)$, from $s=0$ to $s=1$, starting from $z_0 \sim p_{\text{prior}}$.

\textbf{Mean Flow:} The marginal velocity field $v(z_s,s)$ (and thus $v_\theta$) typically contains curvature due to the expectation over $z_s$, even though any given conditional velocity $v_c = (x-e)$ is straight \cite{geng2025meanflowsonestepgenerative}. This means that several ODE integration steps utilizing $v_\theta$ are needed to obtain high-quality samples $z_1$. Mean Flow \cite{geng2025meanflowsonestepgenerative} circumvents this problem by instead modeling the \textit{average velocity} $u(z_s,s,r)$ over the interval $s \rightarrow r$:
\begin{equation}\label{eq:average_velocity}
    u(z_s,s,r) = \frac{1}{r-s} \int_s^r v(z_\tau,\tau) d\tau.
\end{equation}
The estimation of the right-hand side of Eq.~\ref{eq:average_velocity} is intractable. Instead, a tractable training objective is obtained via the \textit{Mean Flow Identity} \cite{geng2025meanflowsonestepgenerative}:
\begin{equation}
    u(z_s,s,r) = v(z_s,s) - (r-s)\frac{d}{ds}u(z_s,s,r).
\end{equation}
By approximating $v(z_s,s)$ by the conditional velocity $(x-e)$ and $u$ with the neural network $u_\theta$, we obtain the training target $u_{\text{tgt}} = (x-e) - (r-s)\frac{d}{ds}u_\theta(z_s,s,r)$, and the Mean Flow loss:
\begin{equation}\label{eq:mf_loss}
    \mathcal{L} = \mathbb{E}_{s,r,x,e}\| u_\theta(z_s,s,r) - \text{sg}(u_{\text{tgt}})\|^2,
\end{equation}
where $\text{sg}(\cdot)$ indicates the stop-gradient operator, and the derivative $\frac{d}{ds}u_\theta$ in the regression target is computed online using Jacobian-vector products. With a trained approximator $u_\theta$, direct sample generation is possible using $z_1 = z_0 + u_\theta(z_0,s=0,r=1)$, where $z_0 \sim p_{\text{prior}}$.

\textbf{Flow-based robotic IL:} Assume that a dataset $\mathcal{D} = \{a_t,o_t\}$ is given, where $a_t$ is the robot action at timestep $t$ and $o_t$ is the corresponding observation of images and/or robot/object states. The dataset $\mathcal{D}$ is collected from an expert policy $\pi_\text{E}$ that we wish to imitate. We follow Diffusion Policy \cite{chi2023diffusion} and define the underlying conditional distribution $p_\text{data}(A_t|O_t)$ induced by the expert, with action sequences $A_t = a_{t-T_o:t+T_p}-1$ and observation sequences $O_t = o_{t-T_o:t}$. The action horizon $T_p$ is the number of future action steps to predict (including the current step, $t$), while the observation horizon $T_o$ is the number of past observations to include as conditioning. Note that the past $T_o$ actions are included in the action sequence $A_t$. The total sequence length is then $T = T_o + T_p$. Given action sequences $A_t$ and Gaussian noise $e \sim \mathcal{N}(0,I)$, define $z_s = A_t^s = sA_t + (1-s)e$ and $v_c = (A_t - e)$. Then IL policies can be constructed using either Flow Matching or Mean Flows. The FM policy is trained by minimizing the loss (adapted from Eq.~\ref{eq:cfm_loss}):
\begin{equation}\label{eq:fm_action_loss}
    \mathbb{E}_{s,A_t,O_t,e} \|v_\theta(A_t^s,O_t,s) - (A_t - e)\|^2.
\end{equation}
With a trained network, robot action sequences $A_t^{s=1}$ are sampled by integrating $\frac{d}{ds}A_t^s = v_\theta(A_t^s,O_t,s)$, starting from $A_t^0 \sim p_{\text{prior}}$, given the most recent observation sequence $O_t$.

For the MF policy, Eq.~\ref{eq:mf_loss} is modified to:
\begin{equation}\label{eq:mf_action_loss}
    \mathcal{L}= \mathbb{E}_{s,r,A_t,O_t,e} \|u_\theta(A_t^s,O_t,s,r) - \text{sg}(u_{\text{tgt}})\|^2,
\end{equation}
where
\begin{equation}\label{eq:u_target_action}
    u_{\text{tgt}} = (A_t - e) - (r-s)\frac{d}{ds}u_\theta(A_t^s,O_t,s,r).
\end{equation}
After training, robot actions can be sampled directly using $A_t^{s=1} = A_t^0 + u_\theta(A_t^0,O_t,0,1)$.

\section{Adaptive Mean Flow}
In this section, we present our proposed Adaptive Mean Flow (AMF).
We leverage Mean Flow's fast, direct action sampling to achieve tighter closed-loop control. A problem to be addressed is that frequent replanning can introduce jittery, inconsistent behavior, as it allows the policy to rapidly switch between behavior modes. As noted in \cite{torne2025ptp} and \cite{black2025realtimeexecutionactionchunking}, the predicted action sequence at the previous step ($A_{t-1}$) can be used as an anchor to steer the new action sequence prediction $A_t$, as long as a subset of the predicted actions from each sequence corresponds to the same timesteps. We take advantage of this, and instead of predicting $A_t$ starting from pure noise $A_t^0 = e$, we start from 
\begin{equation}\label{eq:At-sc}
A_{t^-}^{s} = sA_{t^-} + (1-s)e,
\end{equation}
where 
\begin{equation}\label{eq:action_shift}
    A_{t^-} = \text{Concat}([A_{t-1}]_{1:T-1},[A_{t-1}]_{T-1}).
\end{equation}
$A_{t^-}$ is obtained by removing the first element and copying the last element of $A_{t-1}$, effectively shifting the sequence one step forward. If we write $A_t = A_t(O_t)$ as the current action sequence prediction given the most recent observation sequence, then we can view $A_{t^-} \approx A_t(O_{t-1})$ as the overlapping action sequence given the observation sequence from the previous step. The new prediction $A_t$ becomes
\begin{equation}
A_t^{s=1} = A_{t^-}^{s} + (1-s)u_\theta(A_{t^-}^{s},O_t,s,1)
\end{equation}
As we increase $s$, the relative importance of $A_{t^-}$ and thus $A_{t-1}$ increases, as can be seen from Eqs.~\ref{eq:At-sc} and \ref{eq:action_shift}. Consequently, $A_t$ is likely to remain more similar with $A_{t^-}$.

A drawback of this scheme is that it uses a single value $s$ to specify the noise level uniformly across the sequence. Ideally, we want to keep past and current predictions $a_{t-T_o:t}$ from $A_t$ more closely aligned with $A_{t^-}$, while allowing future predictions $a_{t+1:t+T_p-1}$ more freedom to change with respect to $A_{t^-}$ \cite{black2025realtimeexecutionactionchunking}. To enable this, we follow \cite{chenforcing2024,høeg2024streamingdiffusionpolicyfast} and replace the uniformly perturbed action sequence $A_t^s$ with the variable-noise sequence:
\begin{equation}\label{eq:variable_noise_action_sequence}
    \begin{aligned}
        A_t^S & = [a_{t-T_o}^{s_0},a_{t-T_o+1}^{s_1},...,a_{t+T_p-1}^{s_{T-1}}] \\
        & = [(s_0a_{t-T_o} + (1-s_0)e_0), \\
        & \;\;\;\;\;\;\;...\;,(s_{T-1}a_{t+T_p-1} + (1-s_{T-1})e_{T-1})].
    \end{aligned}
\end{equation}

To enable training and inference with variable noise levels, the scalar start-point $s$ and end-point $r$ are replaced with the vectors $S=s_{0:T-1}$ and $R=r_{0:T-1}$ of length $T$.

\begin{figure*}[t]
    \vspace{-5pt}
    \centering
    \includegraphics[width=.7\textwidth]{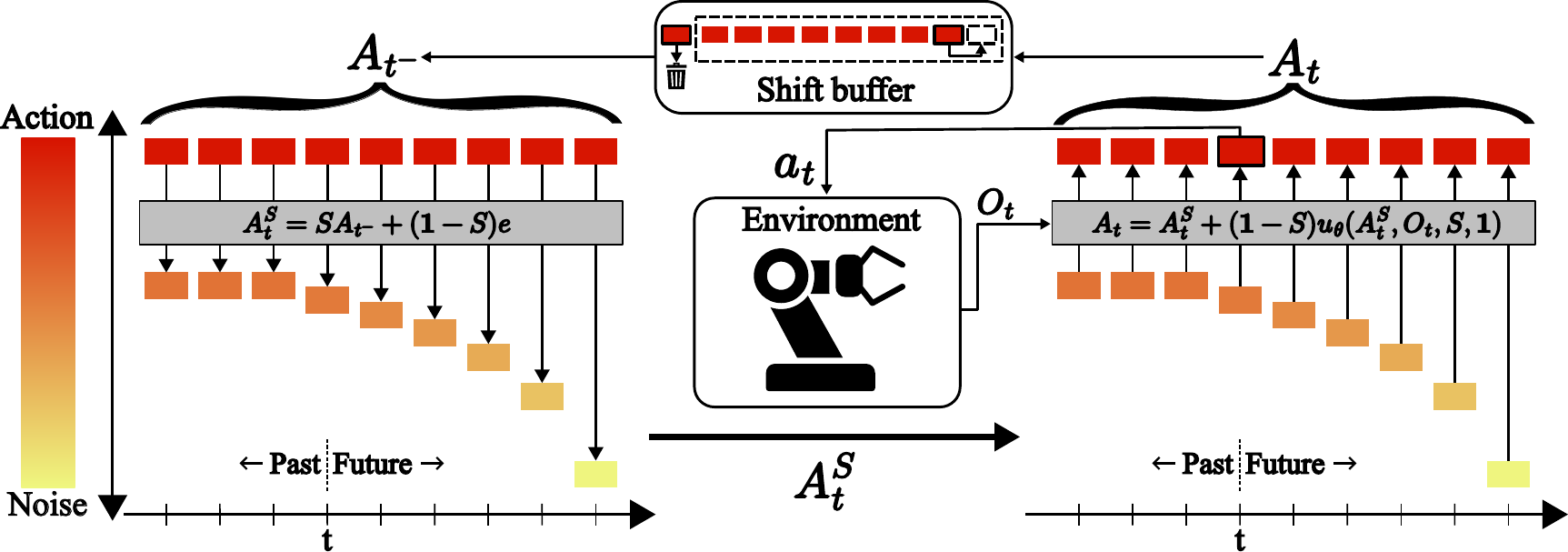}
    \caption{AMF inference. From the left: The previous action sequence prediction $A_{t^-}$ is corrupted with Gaussian noise according to Eq.~\ref{eq:noise_sequence}, producing $A_t^S$. Actions further in the future are more uncertain and, therefore, more corrupted. The Mean Flow network $u_\theta(\cdot)$ predicts the next action sequence $A_t$, using $A_t^S$ and the most recent observation sequence $O_t$. The action $a_t$ corresponds to the current time-step sent to the robot for execution, and $A_t$ is shifted (Eq.~\ref{eq:action_shift}) and reused as $A_{t^-}$ for the next control step.}
    \label{fig:amf_inference}
    \vspace{-10pt}
\end{figure*}

\textbf{Inference with varying noise levels:} During inference, we wish to use a sequence $S$ that is constant for $t-T_o:t-1$, and decreases monotonically for $t:t+T_p-1$ (corresponding to more noise). To this end, we define:
\begin{equation}\label{eq:noise_sequence}
    \begin{aligned}
        s_i & = \begin{cases}
                s_a, & \text{if } i < T_o \\
                s_a \left(\frac{s_b}{s_a}\right)^{\lambda_\gamma(i)}, & \text{if } T_o \leq i < T, \\
            \end{cases}\\
        \lambda_\gamma(i) & = \left (\frac{i-(T_o-1)}{T-T_o} \right )^\gamma, \; \text{for } i \in [T_o,T).
    \end{aligned}
\end{equation}
Here, $s_a$ defines the constant flow index for the initial steps $0:T_o-1$ (corresponding to $t-T_o:t-1$), and $s_b < s_a$ defines the index at the final step $T-1$, while $\gamma$ decides how quickly it approaches $s_b$ in the intermediate steps $T_o:T-2$ (corresponding to $t:t+T_p-2$). The inference procedure is summarized in 
Algorithm~\ref{alg:amf_inference}. In our implementation, we always set $s_b = 0.0$, meaning that the action corresponding to the last prediction step becomes pure noise. We also set $s_a = 0.5$ and $\gamma = 1.5$ for all experiments.

\textbf{Training with varying noise levels:} When going from constant sequence-wide flow intervals $s,r$ to variable intervals $S=s_{0:T-1},R=r_{0:T-1}$, we must decide how to sample $S$ and $R$ during training. The most obvious choice is to sample each $(s_i,r_i)$ independently, with the same sampling schedule as used in the constant $(s,r)$ case. For this ``independent'' regime, we follow \cite{geng2025meanflowsonestepgenerative} and sample pairs $(s_i,r_i)$ using the LogNorm schedule \cite{esserscaling2024}, while making sure $s_i \leq r_i$. This regime ensures high diversity in the sampled noise intervals $S,R$. However, during inference, we intend to use the monotonically decreasing schedule in Eq.~\ref{eq:noise_sequence} for $S$, while setting  $R=1$. Thus, to more closely align the distribution of $S,R$ seen during training with the inference-time distribution, with probability $p_m$ we sample $S$ using Eq.~\ref{eq:noise_sequence}, with $s_a \sim \mathcal{U}[s_{a,\mathrm{min}}, s_{a,\mathrm{max}}]$, $s_b \sim \mathcal{U}[s_{b,\mathrm{min}}, s_{b,\mathrm{max}}]$ and $\gamma \sim \mathcal{U}[\gamma_{\mathrm{min}}, \gamma_{\mathrm{max}}]$, while $R$ is sampled using a constant $r_{0:T-1} = r$, with $r$ drawn from a LogNorm with expectation close to $1$. The training pipeline is presented in Algorithm~\ref{alg:amf_training}.

\begin{figure}[t]
    \centering
    \includegraphics[width=.5\textwidth]{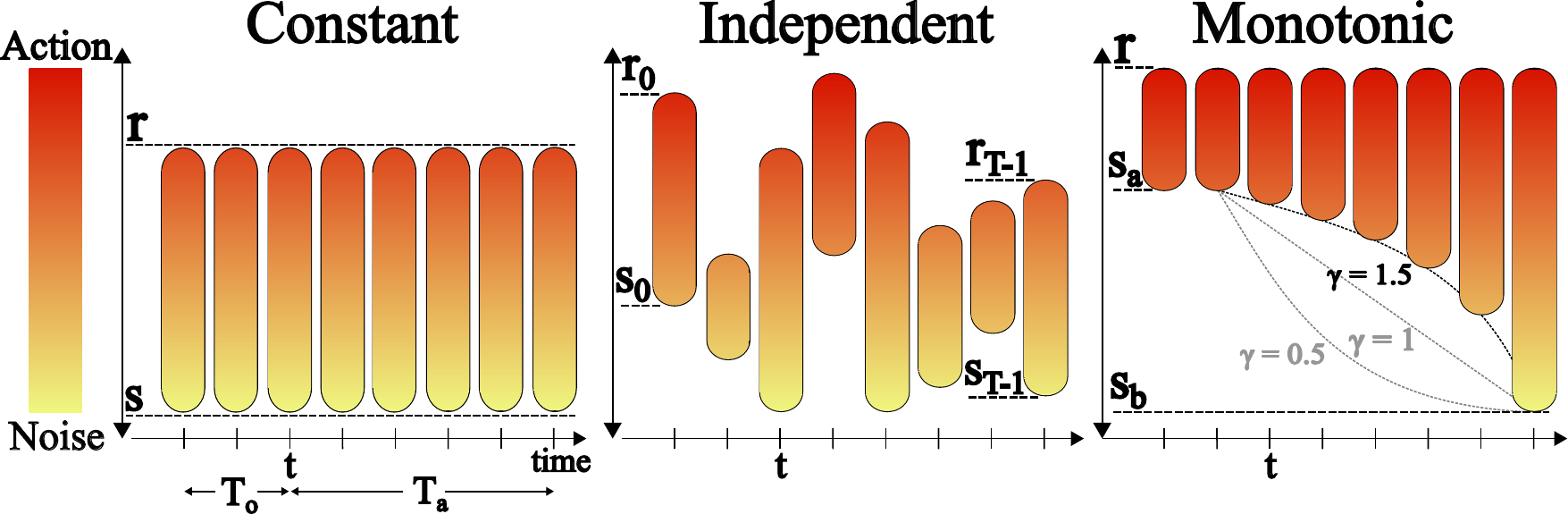}
    \caption{Flow interval sampling for an example action sequence with $T_o=2$ and $T_p=6$. Instead of using the same values $s,r$ across the action sequence (left), AMF uses independent intervals $s_i,r_i$. During training, we use a mix of independently sampled intervals (middle) and monotonically increasing intervals (right).}
    \label{fig:mean_flow_interval_sampling}
    \vspace{-15pt}
\end{figure}

\textbf{Conditioning on flow index intervals $S$ and $R$:} When going from scalar flow intervals $(s,r)$ to interval sequences $(S,R)$, an important design choice is how to condition the network $u_\theta$ on $S$ and $R$.  We conduct experiments using three different conditioning variants: concatenation (\textbf{Concat}), addition (\textbf{Add}), and feature-wise Adaptive LayerNorm-Zero (\textbf{F-AdaLN-Z}) \cite{peebles2023dit}. \textbf{Concat} simply involves appending the values $(s_i,r_i)$ to the corresponding noisy action $a_i^s$. With \textbf{Add}, we first project $s_i$ and $r_i$ to the embedding dimension and add them to the action tokens. \textbf{F-AdaLN-Z} uses $s_i$ and $r_i$ to parametrize zero-initialized adaptive normalization layers applied individually to each action token (feature).

\textbf{Predicting past actions:} The observation horizon $T_o$ denotes the number of past observation steps that is used as input when predicting an action sequence. As shown in Figure~\ref{fig:amf_inference}, the policy also predicts the \textit{past actions} $a_{t-T_o:t-1}$. As shown in \cite{torne2025ptp}, predicting past robot actions and aligning these predictions with the true past robot trajectory improves temporal action consistency and overall performance. Since AMF initializes the current predicted sequence from the previous, including partially noised actions from the past implicitly encourages consistent policy behavior. 

% \bigskip
\begin{figure}[t]
\vspace{-15pt}
\begin{minipage}{0.95\linewidth}
\begin{algorithm}[H] % Or use [htbp] for more flexible placement
\small
\caption{\textbf{AMF Inference}}
\label{alg:amf_inference}
\begin{algorithmic}[1]
\REQUIRE Mean Flow Network $u_\theta(\cdot)$
\WHILE{not terminated}
    \STATE $O_t \gets T_o$ most recent observations
    \IF{$t == 0$}
    \STATE $S  \gets \boldsymbol{0}$
    \STATE $A_{t^-} \gets \boldsymbol{0}$
    \ELSE
    \STATE Construct $S$ with fixed $s_a,s_b$ and $\gamma$, (Eq.~\ref{eq:noise_sequence})
    \STATE Construct $A_{t^-}$ with $A_{t-1}$, 
    (Eq.~\ref{eq:action_shift})
    \ENDIF
    \STATE Sample $e \sim \mathcal{N}(0,I)$
    \STATE Construct $A_t^S$ with $A_{t^-}$, $e$ and $S$, (Eq.~\ref{eq:variable_noise_action_sequence})
    \STATE $A_t \gets A_t^S + (\textbf{1}-S)u_\theta(A_t^S,O_t,S,\textbf{1})$
\ENDWHILE
\end{algorithmic}
\end{algorithm}
\end{minipage}
\vspace{-20pt}
\end{figure}

% \bigskip
\begin{figure}[t]
\vspace{-10pt}
\begin{minipage}{0.95\linewidth}
\begin{algorithm}[H] % Or use [htbp] for more flexible placement
\small
\caption{\textbf{AMF Training}}
\label{alg:amf_training}
\begin{algorithmic}[1]
\REQUIRE Dataset $\mathcal{D}$, Mean Flow Network $u_\theta(\cdot)$
\WHILE{not converged}
    \STATE Sample $(A_t, O_t) \sim \mathcal{D}$
    \STATE Sample $p \sim \mathcal{U}[0,1]$
    \IF{$p \leq p_m$}
    \STATE Sample $s_a \sim \mathcal{U}[s_{a,\mathrm{min}}, s_{a,\mathrm{max}}],$
    \STATE $\quad s_b\sim \mathcal{U}[s_{b,\mathrm{min}}, 
    s_{b,\mathrm{max}}],\gamma \sim \mathcal{U}[\gamma_{\mathrm{min}}, \gamma_{\mathrm{max}}]$
    \STATE Construct $S$ with $s_a,s_b$ and $\gamma$, 
    (Eq.~\ref{eq:noise_sequence})
    \STATE Sample $R = r \sim \text{LogNorm}(\mu=\mu_h,\sigma=1.0)$
    \ELSE
    \STATE Sample $s_i,r_i \sim \text{LogNorm}(\mu=\mu_{l},\sigma=1.0)$
    \STATE Ensure $s_i \leq r_i$
    \ENDIF
    \STATE Sample $e \sim \mathcal{N}(0,I)$
    \STATE Construct $A_t^S$ with $A_t$, $e$ and $S$, (Eq.~\ref{eq:variable_noise_action_sequence})
    \STATE $\hat{u} \gets u_\theta(A_t^S,O_t,S,R)$
    \STATE Construct $u_{\text{tgt}}$ with $A_t,e,\hat{u},S$ and $R$, 
    (Eq.~\ref{eq:u_target_action})
    \STATE $\theta \gets \theta - \nabla_\theta\mathcal{L}(\hat{u},u_{\text{tgt}})$
\ENDWHILE
\end{algorithmic}
\end{algorithm}
\end{minipage}
\vspace{-10pt}
\end{figure}

\section{Simulation Experiments}
In this section, we present a comprehensive set of simulated experiments to evaluate the overall efficacy of our proposed method and investigate key design choices.

\subsection{Simulation environments}
\textbf{Metaworld:} To assess each method's capability to perform varied, precise, and smooth robot manipulation, we extract 8 challenging tasks from the widely used Metaworld object manipulation benchmark \cite{mclean2025metaworld}. For each task, we collect 100 expert demonstrations using pre-trained RL policies included in the official repository of \cite{mclean2025metaworld}. The action space consists of 3DOF end-effector velocities and gripper open/close. To test policy responsiveness and robustness, we conduct additional experiments in which correlated noise is added to the robot's actions, requiring policies to adjust quickly to end-effector perturbations in order to solve the tasks.

\textbf{Push-T:} Push-T is a 2D manipulation task where the agent must carefully align a T-shaped object within a goal region. The dataset contains 90 highly multimodal demonstrations collected via human teleoperation, so smoothness and temporal mode consistency are important for solving the task.

\textbf{Dynamic Block Push (Dyn-BP)}: This is a dynamic and multimodal 2D manipulation where the agent must push a red and a green block into either of 2 goal zones that are drifting randomly within the environment. The dataset contains 1000 episodes.

\textbf{D3IL:} Lastly we include the Stack-1 and Stack-2 tasks from the D3IL benchmark \cite{jia2024towards}. D3IL tasks contain multiple solution strategies and the corresponding datasets consist of diverse, human-collected demonstrations. Stack-1 and Stack-2 evaluate the agent's ability to successfully stack 1 block or 2 blocks, respectively. Dataset size: 1000 episodes.

\subsection{Baselines}

We compare AMF against a varied set of baseline policies. All presented policies use a transformer architecture based on \cite{chi2023diffusion} as the policy network. Noisy robot actions $a^s$ represent the transformer input tokens, while observations $o$ and flow indices $s$ (and optionally $r$) are projected to the token dimension and incorporated using cross-attention.  The transformer network has 8 layers, 4 heads, and an embedding size of 256. All policies use the past $T_o=9$ observation frames and predict $T_p=10$ future actions. For the chunking variants, we set the chunk length to 8. With a control frequency of 10Hz, this gives an open-loop horizon of 0.8s, which has been shown to work well for quasi-static tasks \cite{chi2023diffusion}.

\textbf{FM-8}, \textbf{FM-1} and \textbf{FM-RTC} predict robot actions using Flow Matching \cite{lipman2023flow}, and are trained using Eq.~\ref{eq:fm_action_loss}. FM-8 uses a chunk length of $T_a = 8$ (replans every 8th step), while FM-1 uses single-step action execution. FM-RTC uses Real-Time Chunking \cite{black2025realtimeexecutionactionchunking} at inference time, an algorithm that leverages past action sequence predictions to improve consistency across subsequent model predictions. \textbf{AFM} is equivalent to AMF but with Flow Matching instead of Mean Flow. AFM uses the same architecture as AMF. All FM-based methods use 10 denoising iterations.

\textbf{MF-8} and \textbf{MF-1} use Mean Flow \cite{geng2025meanflowsonestepgenerative} (Eq.~\ref{eq:mf_action_loss}), and predict action sequences directly from Gaussian noise in a single denoising step. MF-8 and MF-1 use chunk lengths of 8 and 1, respectively. \textbf{MF-WarmStart} (\textbf{MF-WS}) reuses past trajectories similar to AMF, but with a constant noise-level across the trajectory instead of increasing noise. MF-WS is evaluated with $T_a=1$ and a corruption level of $s=0.5$.

\textbf{SDP} \cite{høeg2024streamingdiffusionpolicyfast} uses variable-noise diffusion to reduce the number of denoising iterations needed to predict action sequences. We implemented SDP using the same transformer architecture and time-step conditioning (AdaLNZero) as AMF. \textbf{Consistency Policy (CP)} \cite{prasadconsistency} distills a diffusion-based action policy into a student model using consistency-preserving objective \cite{pmlr-v202-song23a}, allowing inference with a single denoising step. We evaluate both SDP and CTM with $T_a=1$.

\subsection{Results}

\begin{table*}[t]
\centering
\caption{Main simulation results. For Push-T and Metaworld, the models are trained for 500 epochs, while for Dyn-BP and D3IL (Stack-1 and Stack-2) the models are trained for 200 and 90 epochs, respectively. We use a batch size of 256, and model checkpoints are collected every 10th epoch. For Push-T and Dyn-BP, the score is the maximum goal overlap ratio achieved during an episode, whereas for Metaworld and D3IL, the score is 1 for success and 0 otherwise. Each model rollout consists of 50 tests, and we report averaged metrics over the last three model checkpoints, across three seeds. Standard deviation is shown in parentheses.}
\setlength{\tabcolsep}{1.pt}
\renewcommand{\arraystretch}{1.1}
\resizebox{\textwidth}{!}{\begin{tabular}{lccccccccccccc}
\toprule
\taskimg{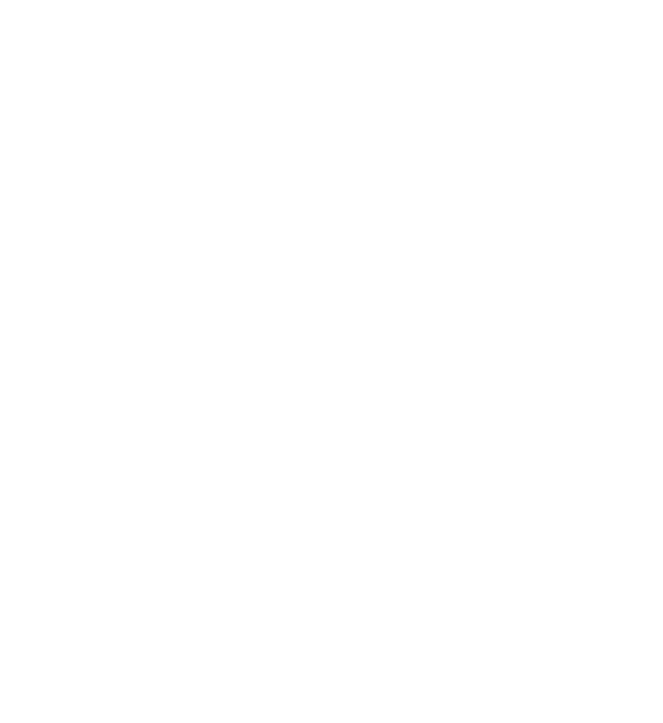}{}
& \taskimg{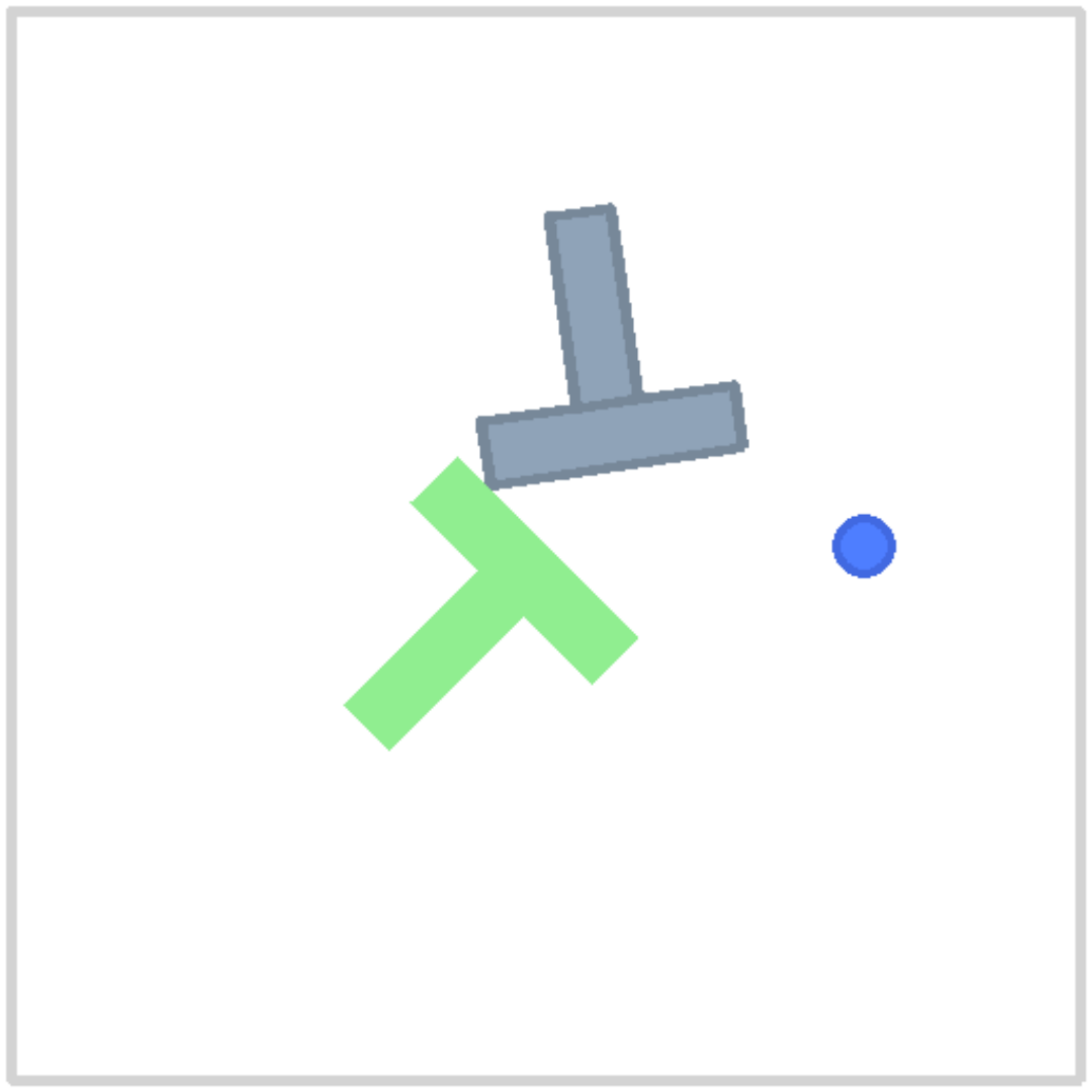}{Push-T}
& \taskimg{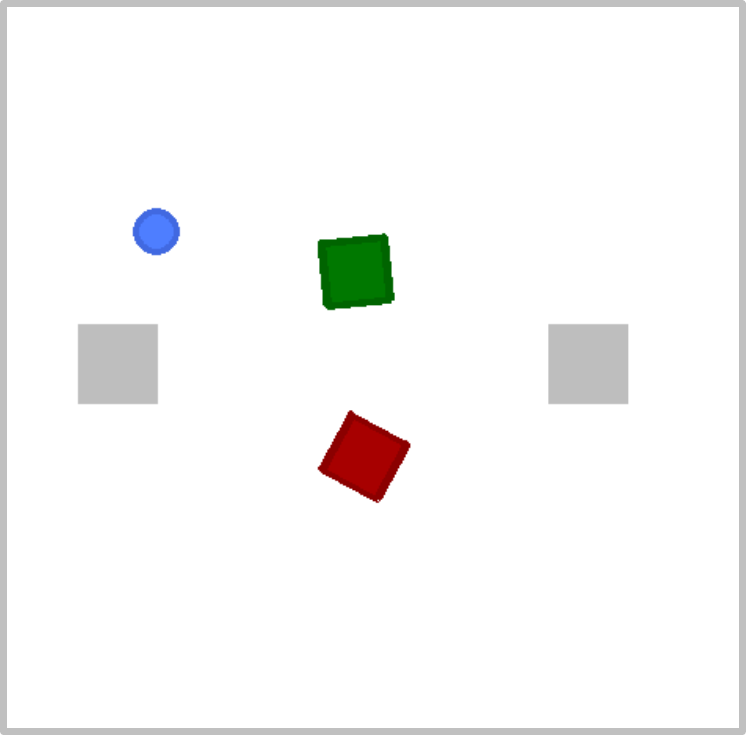}{Dyn-BP}
& \taskimg{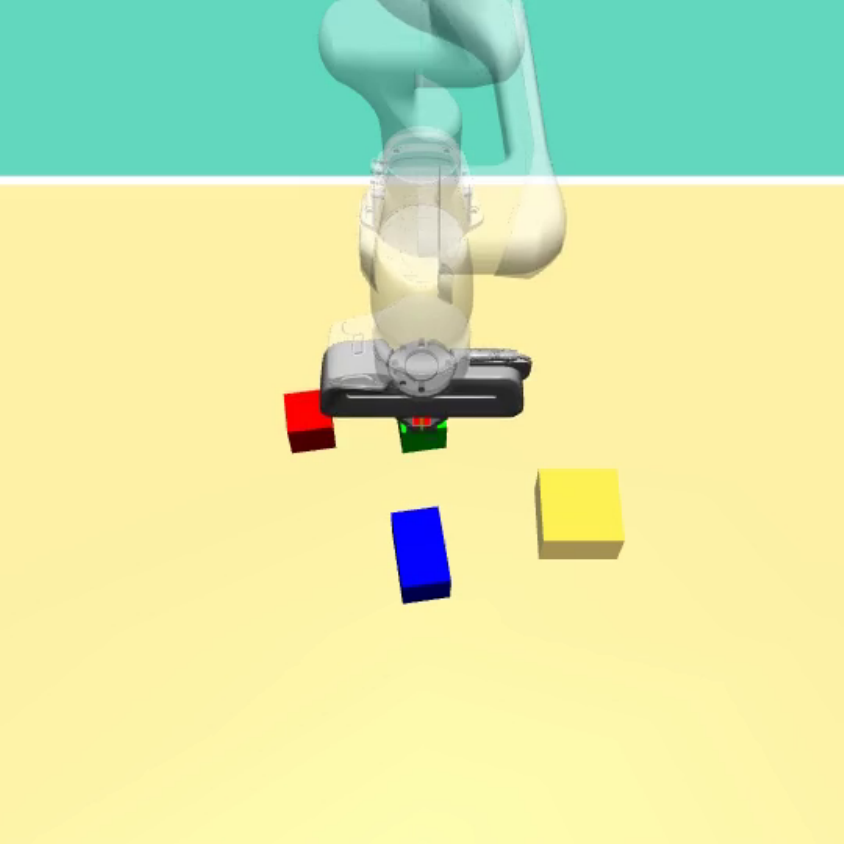}{Stack-1}
& \taskimg{figs/d3il_stacking_thumbnail.pdf}{Stack-2}
& \taskimg{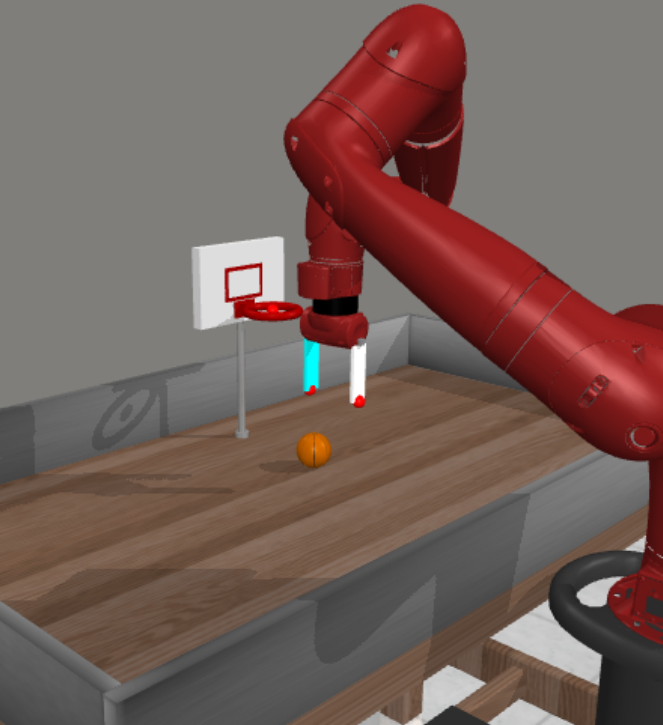}{Basketball}
& \taskimg{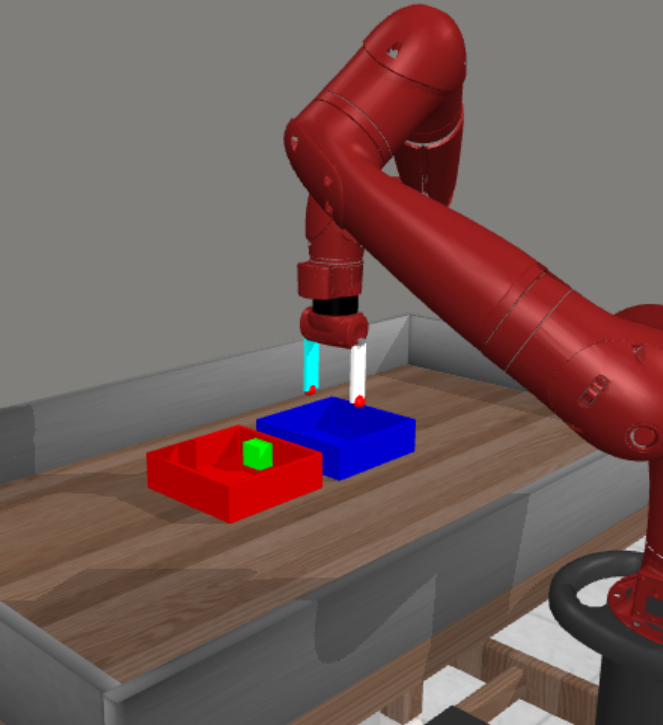}{Bin-picking}
& \taskimg{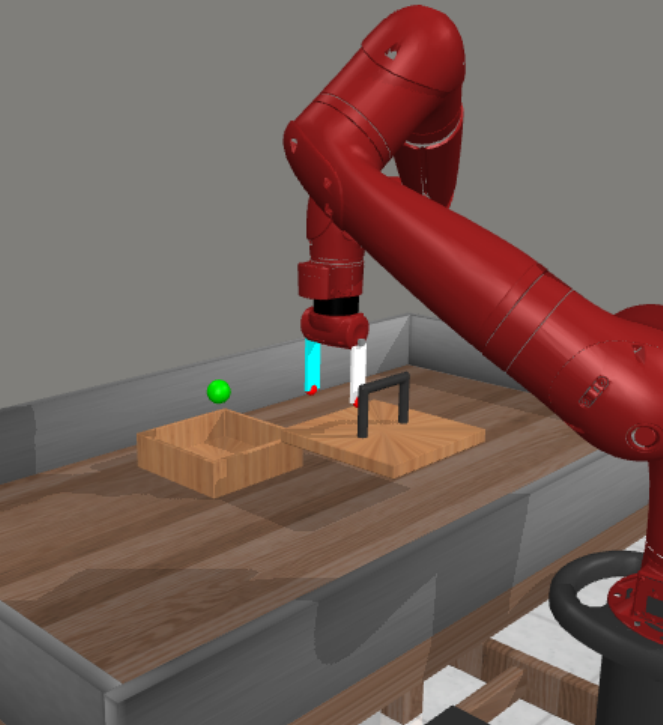}{Box-close}
& \taskimg{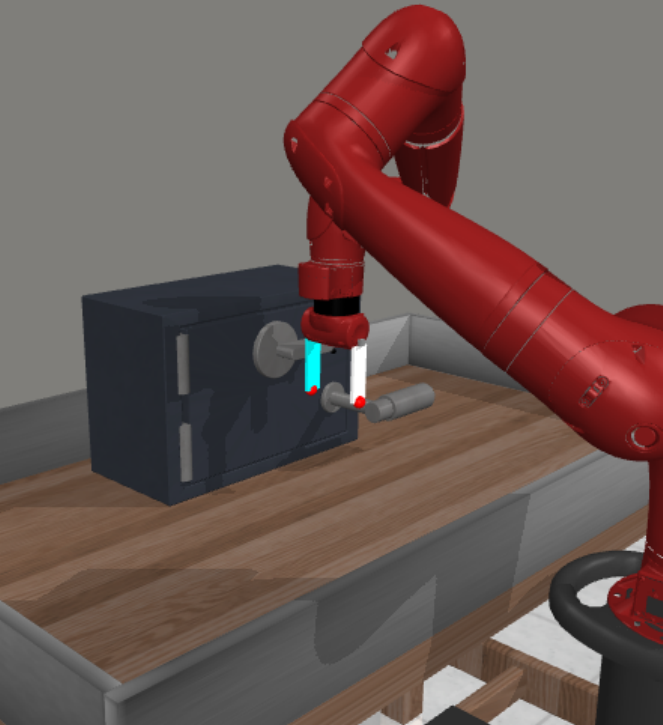}{Door-lock}
& \taskimg{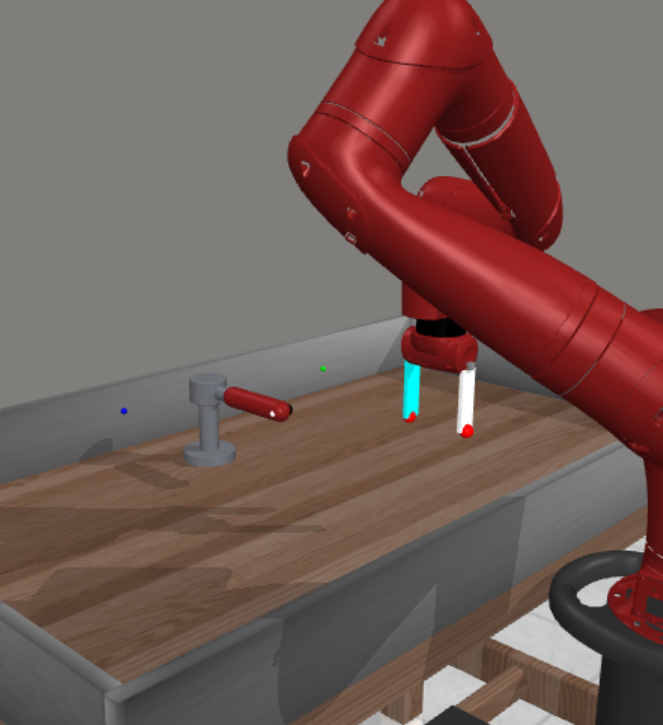}{Faucet-close}
& \taskimg{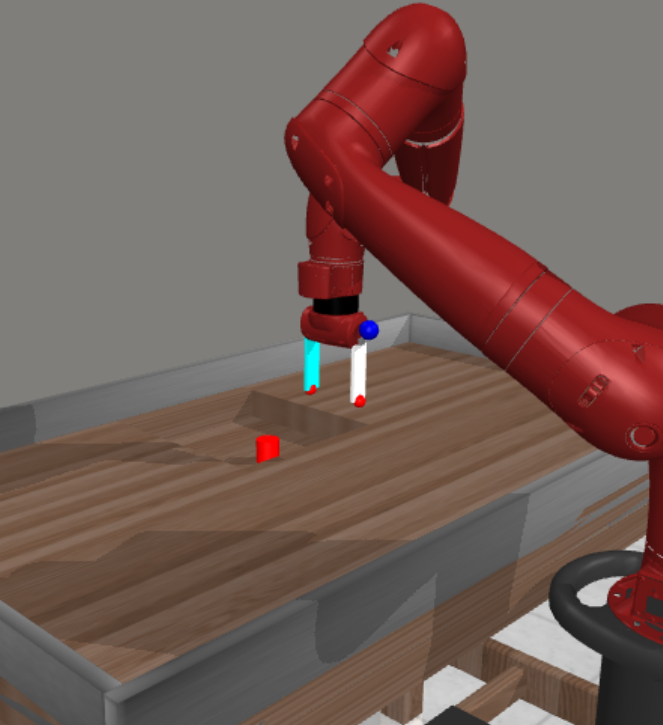}{Pick-out-of-hole}
& \taskimg{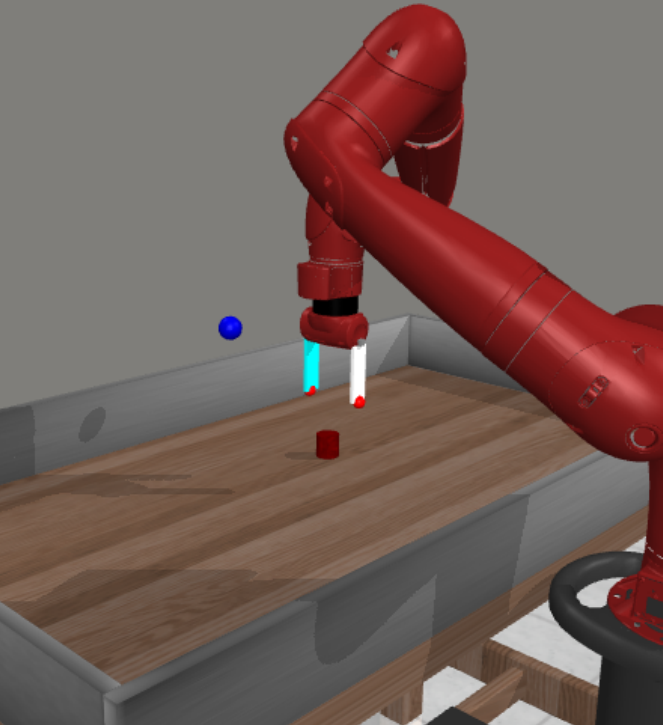}{Pick-place}
& \taskimg{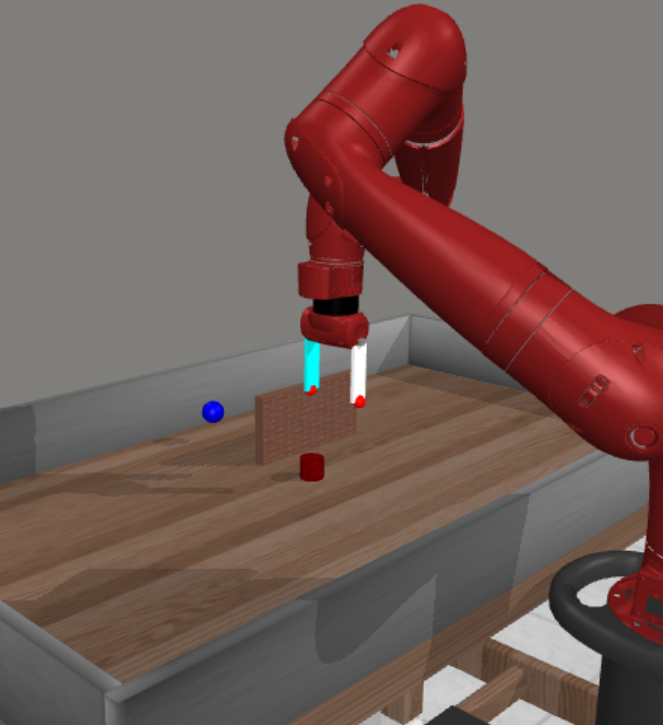}{Pick-place-wall}
& \taskimg{figs/image_placeholder.pdf}{}\\
\midrule
Method & Overlap & Overlap & SR & SR & SR & SR & SR & SR & SR & SR & SR & SR & \textbf{Avg}\\
\midrule
FM-8          & 0.73 (0.05) & 0.21 (0.01) & \textbf{0.67} (0.07) & 0.24 (0.07) & 0.76 (0.10) & 0.85 (0.07) & 0.31 (0.03) & 0.86 (0.07) & 0.95 (0.03) & 0.59 (0.05) & 0.90 (0.07) & 0.92 (0.04)  & 0.67\\
FM-1  & 0.83 (0.02) & 0.31 (0.03) & 0.43 (0.05) & 0.08 (0.04) & 0.95 (0.05) & 0.88 (0.02) & 0.59 (0.12) & 0.94 (0.03) & 0.99 (0.01) & 0.50 (0.22) & 0.83 (0.06) & 0.92 (0.04) & 0.69\\
FM-RTC  & 0.79 (0.01) & 0.30 (0.02) & 0.46 (0.12) & 0.08 (0.06) & 0.92 (0.07) & 0.90 (0.05) & 0.57 (0.13) & 0.98 (0.03) & 0.99 (0.01) & 0.50 (0.18) & 0.81 (0.06) & 0.90 (0.04) & 0.68\\
AFM  & 0.82 (0.05) & 0.44 (0.04) & 0.63 (0.05) & 0.24 (0.04) & 0.93 (0.01) & 0.79 (0.32) & \textbf{0.92} (0.04) & \textbf{1.00} (0.00) & \textbf{1.00} (0.00) & 0.63 (0.09) & 0.84 (0.05) & 0.93 (0.03) & 0.76\\
MF-8  & 0.65 (0.04) & 0.21 (0.02) & 0.20 (0.24) & 0.06 (0.07) & 0.20 (0.12) & 0.48 (0.20) & 0.63 (0.14) & 0.87 (0.14) & 0.96 (0.06) & 0.08 (0.03) & \textbf{0.93} (0.04) & 0.90 (0.05) & 0.51\\
MF-1  & 0.75 (0.02) & 0.34 (0.06) & 0.03 (0.08) & 0.01 (0.02) & 0.41 (0.10) & 0.59 (0.14) & 0.51 (0.08) & 0.96 (0.06) & \textbf{1.00} (0.00) & 0.17 (0.09) & 0.78 (0.07) & 0.84 (0.05) & 0.53\\
MF-WS  & 0.81 (0.04) & 0.30 (0.04) & 0.12 (0.14) & 0.01 (0.02) & 0.92 (0.03) & 0.86 (0.03) & 0.79 (0.12) & 0.97 (0.05) & \textbf{1.00} (0.00) & 0.33 (0.04) & 0.87 (0.02) & \textbf{0.95} (0.04) & 0.66\\
SDP  & 0.90 (0.03) & 0.39 (0.00) & 0.56 (0.10) & 0.21 (0.08) & 0.76 (0.06) & 0.67 (0.18) & \textbf{0.92} (0.04) & \textbf{1.00} (0.00) & 0.18 (0.13) & 0.90 (0.05) & 0.79 (0.16) & 0.53 (0.14) & 0.65\\
CP  & 0.34 (0.33) & 0.03 (0.00) & 0.00 (0.00) & 0.00 (0.00) & 0.67 (0.25) & 0.01 (0.01) & 0.2 (0.1) & 0.68 (0.44) & 0.56 (0.38) & 0.02 (0.00) & 0.83 (0.12) & 0.81 (0.10) & 0.35\\
AMF (ours)    & \textbf{0.92} (0.02) & \textbf{0.50} (0.05) & 0.58 (0.08) & \textbf{0.27} (0.08) & \textbf{0.97} (0.02) & \textbf{1.00} (0.00) & 0.86 (0.07) & \textbf{1.00} (0.00) & \textbf{1.00} (0.00) & \textbf{0.92} (0.05) & 0.83 (0.04) & 0.93 (0.04) & \textbf{0.82}\\
\bottomrule
\end{tabular}}
\label{tab:main_sim_results}
\vspace{-10pt}
\end{table*}

Table~\ref{tab:main_sim_results} summarizes the main simulation results. AMF performs stably across all tasks and achieves a 6\% relative improvement over the best-performing baseline, AFM. The Mean Flow baselines MF-8 and MF-1 perform worse than the Flow Matching baselines, indicating that naively swapping Flow Matching for Mean Flow, while enabling much faster inference, reduces model prediction quality. Note that inference delay does not impact the simulation environments, so we defer this analysis to Section~\ref{sec:real_world}. AMF (and AFM) performs significantly better than SDP, even though both methods use adaptive noise schedules and the same architecture. AMF fully denoises the predicted trajectory at every step and subsequently corrupts it with fresh noise (according to the adaptive schedule). SDP, by contrast, partially denoises each action in the trajectory at each step, appending fresh noise only at the end (i.e., the farthest future point) of the action sequence. We hypothesize that the added variation provided by the AMF/AFM approach leads to better performance compared with SDP. 
CP performs relatively poorly, and the training of the student policy was unstable for some tasks. We used the same transformer-based architecture for all baselines, which differs from the Unet architecture used in \cite{prasadconsistency}. However, training instability is a known drawback of consistency models \cite{kim2026stabilizingconsistencytrainingflow}.

Metaworld tasks with the largest performance discrepancies, such as Basketball, Box-close, and Pick-out-of-hole, generally require very precise manipulation to succeed. Indeed, we observe that AMF produces smooth and accurate robot trajectories. Push-T, on the other hand, also requires occasional trajectory ``switches'' to solve reliably, e.g., switching from pushing on the side of the block to pushing from the top to align it vertically. If the policy is overly biased towards smooth motion, it is prone to getting stuck in a single task phase and eventually stalling. However, this is not the case for AMF, which we attribute to the increased corruption applied to future action predictions, providing sufficient flexibility to switch trajectories when needed.

\textbf{Smoothness:} In Table \ref{tab:smoothness}, we report trajectory smoothness estimates on the Metaworld action trajectories computed using third-order finite differences. AMF exhibits considerably less jerkiness compared with MF-1 and FM-1, and also compared with the constant-noise trajectory reuse baseline (MF-WS). We also see that the Mean Flow baseline MF-1 is significantly less smooth than its Flow Matching counterpart, which helps explain the performance difference between these two baselines. 

\begin{figure}[t]
    \centering
    \includegraphics[width=0.40\textwidth]{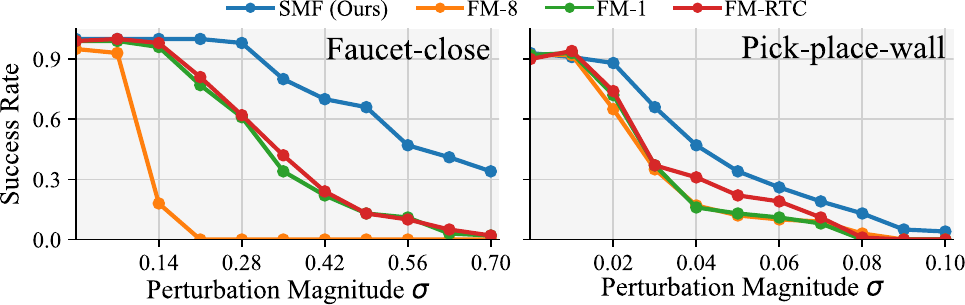}
    \caption{Success rate on 2 Metaworld tasks with increasing perturbations added to the end-effector. Each policy obtains a close to 100\% success rate without perturbation; however, the performance of AMF (blue curve) consistently degrades the least as the perturbation magnitude $\sigma$ (and thus the task difficulty) increases.}
    \label{fig:ac_perturb_ablation}
    \vspace{-5pt}
\end{figure}

\textbf{Robustness to perturbation:} One of the main reasons for using single-step (i.e., fully closed-loop) action execution as opposed to action chunking is to increase robustness to environmental disturbances. In the Dyn-BP task, we see a clear trend where the chunking policies FM-8 and MF-8 perform significantly worse than their single-step variants, which indicates that reactivity is indeed needed to align the blocks with the moving goal zones. To further analyze this, we augment Metaworld with dynamic robot perturbations. The perturbations are implemented as temporally correlated Ornstein-Uhlenbeck noise
\begin{equation}\label{eq:action_perturbation}
    \epsilon_t = \rho \epsilon_{t-1} + \sqrt{1-\rho^2}\sigma w_t, \; w_t \sim \mathcal{N}(0,I),
\end{equation}
which is added to the robot actions as $\tilde{a}_t = a_t + \epsilon_t$. $\rho$ is set to a fixed value, so the perturbation strength is tuned by varying $\sigma$. Figure~\ref{fig:ac_perturb_ablation} compares AMF to selected baselines on two Metaworld tasks, for a wide range of perturbation magnitudes. AMF consistently shows the greatest robustness. The chunking baseline FM-8 performs the worst on average, consistent with the intuition that open-loop execution makes policies less robust to dynamic disturbances. This is particularly evident on the \textit{Faucet-close} task.

\textbf{Design ablations:} Table~\ref{tab:design_ablation} compares the performance of key design variations on Push-T and Metaworld. The best performance is achieved using feature-wise Adaptive LayerNorm-Zero conditioning (\textbf{F-AdaLN-Z}) \cite{peebles2023dit} for the flow intervals $(S,R)$. With this formulation, the values for $s_i$ and $r_i$ are incorporated into the corresponding action tokens in every transformer block, permitting more fine-grained adaptation compared with the \textbf{Concat} and \textbf{Add} variants, which only apply the conditioning once at the input to the transformer. We also see that setting $p_m = 0.5$, i.e., training the model with a mix of sequences with independently sampled flow intervals and sequences with monotonically increasing intervals (Eq.~\ref{eq:noise_sequence}) yields the best performance, indicating that exposing the policy to both corruption patterns during training is beneficial. We also tested a variant of AMF that removes the conditioning on the end-points $R$, since this sequence is always set to $1$ during inference. This variant performs almost as well as the base AMF, indicating that strong performance is achievable even without explicit end-point conditioning.

\begin{figure}[]
    \centering
    \includegraphics[width=0.25\textwidth]{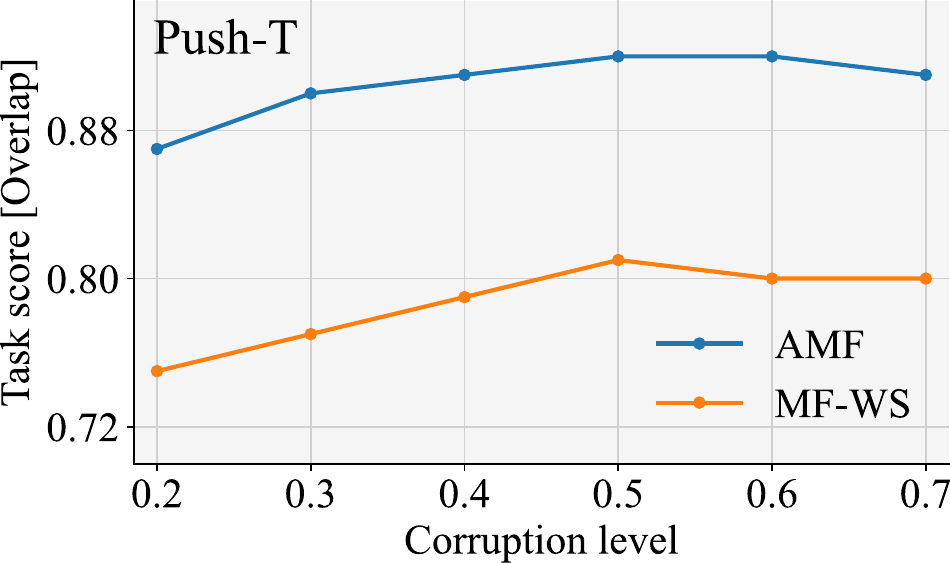}
    \caption{Task score on Push-T for AMF and MF-WS with varying levels of noise corruption. For AMF, the value on the x-axis corresponds to the starting noise $s_{a}$, while $s_b=1.0$ and $\gamma = 1.5$. For MF-WS, the value corresponds to the constant noise corruption level.}
    \label{fig:amf_mfws_corruption_ablation}
    \vspace{-5pt}
\end{figure}

Figure \ref{fig:amf_mfws_corruption_ablation} shows performance on the Push-T task for AMF and MF-WS, with varying levels of noise corruption. Performance remains relatively stable across this range.

\begin{table}[]
    \centering
    \caption{Trajectory smoothness analysis on Metaworld. We quantify action trajectory smoothness using the Euclidean norm of the third finite difference of consecutive Cartesian actions. We report its mean and maximum over evaluation trajectories, where lower values indicate smoother action sequences. The mean and maximum is computed for each task, then averaged across all tasks.}
    \begin{tabular}[width=.48\textwidth]{lcc}
    \toprule
       Method  & Mean & Max\\
       \midrule
        FM-1 & 1.16 & 26.67 \\
        MF-1 & 2.19 & 29.69 \\
        MF-WS & 1.33 & 23.63\\
        AMF (Ours) & 0.98 & 15.96\\
        \bottomrule
        
    \end{tabular}
    \label{tab:smoothness}
    \vspace{-5pt}
\end{table}

\begin{table}[]
    \centering
    \caption{Performance impact from selected design variations. The type of conditioning used for the flow intervals $(S,R)$ is denoted in bold. $p_m$ is the probability of sampling interval sequences $(S,R)$ using the monotone schedule (Eq.~\ref{eq:noise_sequence}), leaving $(1-p_m)$ probability of sampling each $(s_i,r_i)$ independently.}
    \begin{tabular}[width=.48\textwidth]{lcc}
    \toprule
       Ablation  & Push-T (Overlap) & Metaworld Avg. (SR)\\
       \midrule
       \textbf{F-AdaLN-Z} / $p_m = 0.5$ & \textbf{0.92} (0.02) &  \textbf{0.95}\\
        \textbf{Concat} / $p_m = 0.5$ & 0.86 (0.01) & 0.90 \\
        \textbf{Add} / $p_m = 0.5$ & 0.85 (0.00) & 0.92\\
        \textbf{F-AdaLN-Z} / $p_m = 0.0$ & 0.89 (0.02) & 0.93 \\
        \textbf{F-AdaLN-Z} / $p_m = 1.0$ & 0.88 (0.01) &  0.77\\
        $S$-conditioning only & 0.9 (0.02) & 0.93 \\ 
        \bottomrule
        
    \end{tabular}
    \label{tab:design_ablation}
    \vspace{-5pt}
\end{table}

\section{Real-World Evaluation}\label{sec:real_world}
\begin{figure}[t]
    \centering
    \includegraphics[width=0.48\textwidth]{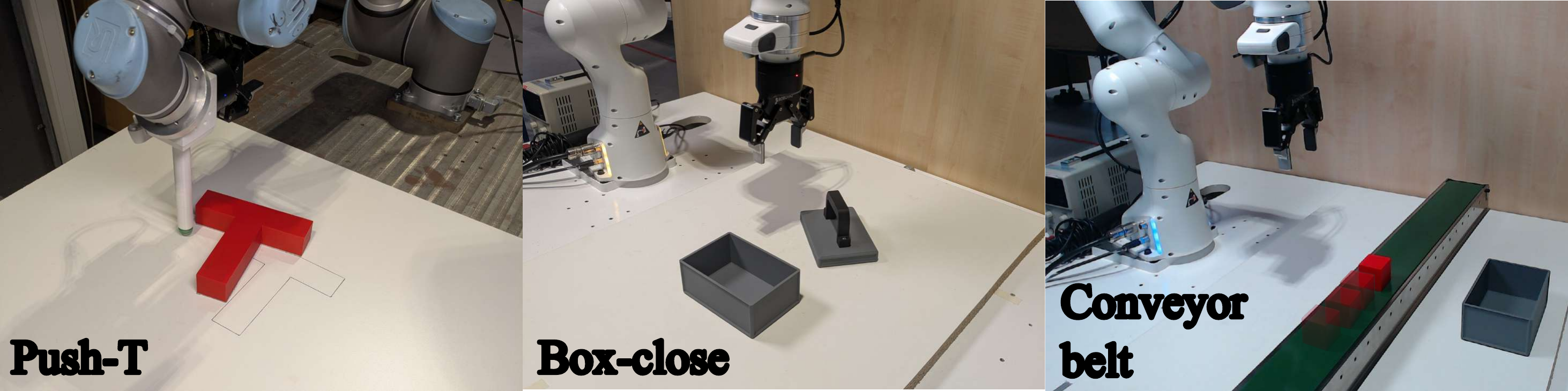}
    \caption{Real-world manipulation tasks.}
    \label{fig:real_tasks}
    %\vspace{-7pt}
\end{figure}

We conduct real-world experiments on three challenging robot manipulation tasks, as visualized in Figure~\ref{fig:real_tasks}. 

\textbf{Real-world Push-T:} This is a real-world implementation of the Push-T task \cite{chi2023diffusion}. The starting position and orientation of the T-shaped block is randomized, and the policies must use a cylindrical rod attached to the end-effector to push the block into the marked goal location. As in the simulated version, the dataset is highly multimodal and contains trajectories collected with different solution strategies.

\textbf{Box-close:} In this task, a lid with a handle must be picked up and accurately placed on the corresponding box. The starting orientations and positions of both the lid and box are randomized, and correctly placing the lid requires a high degree of precision.

\textbf{Conveyor Belt:} A miniature conveyor belt of length 90~cm and width 10 cm is placed in front of the robot, and the task is to intercept a red block moving along the belt and place it in the container. The belt speed varies, so the agent must adapt in real time to successfully pick up the block.

The Real-world Push-T task is implemented using a Universal Robots UR10 manipulator, and the action space consists of 2D delta position references in the $xy$-plane. The Box-close and Conveyor Belt tasks use a Franka Emika Panda setup with a 5-dimensional action space comprising $xyz$-translation, yaw rotation, and gripper open/close. For all tasks, the observations consist of end-effector poses and RGB frames from two RealSense D435 cameras, one mounted on the robot wrist and the other on a camera stand overlooking the workspace. The frames are processed by a pair of ResNet-18 image encoders \cite{he2016deep} trained end-to-end alongside the robot policies. All demonstrations are conducted with a control frequency of 10 Hz. Model inference was conducted locally on a single Nvidia RTX 3090 GPU.

\subsection{Results}

Results for the real Push-T and Box-close tasks are shown in Table~\ref{tab:real_results}. AMF performs best overall and significantly outperforms the baselines on the real Push-T task. Visually, both FM-8 and FM-RTC were noticeably less smooth than AMF, and a typical failure mode for the baselines was abrupt movements that misaligned the T-block in the Push-T case or resulted in an incorrect grip on the Box-close task. A likely contributor is inference latency, which is reported in the rightmost column of Table~\ref{tab:real_results}. The RTC inference algorithm \cite{black2025realtimeexecutionactionchunking} is designed to increase smoothness, but comes at the cost of even slower inference compared with vanilla FM. AMF combines fast inference speed and the smoothing effect of reusing previous predictions to minimize jerkiness. AMF would, however, sometimes become ``stuck'' before fully completing the tasks. For example, almost all AMF failures on the Box-close task were due to not releasing the grip on the lid after aligning it with the box.

\begin{table}[]
    \centering
    \caption{Results on static real-world tasks. Each policy is trained for 600 epochs and evaluated on 20 random initial configurations for each task. For Push-T, the policies receive a score of 1 for successful alignment and 0 otherwise. For Box-close, correctly picking the lid yields 0.5 score, while correctly placing it yields an additional 0.5.}
    \begin{tabular}[width=.48\textwidth]{lcccc}
    \toprule
       Method  & Real Push-T & Real Box-close & Inference latency\\
       \midrule
       FM-8 & 0.35 &  0.48 & 54 ms\\
       FM-RTC & 0.35 &  0.43 & 79 ms\\
       AMF  (Ours) & \textbf{0.60} &  \textbf{0.58} & 14 ms\\
       \bottomrule
        
    \end{tabular}
    \label{tab:real_results}
    \vspace{-10pt}
\end{table}

\begin{figure}[t]
    \centering
    \includegraphics[width=0.37\textwidth]{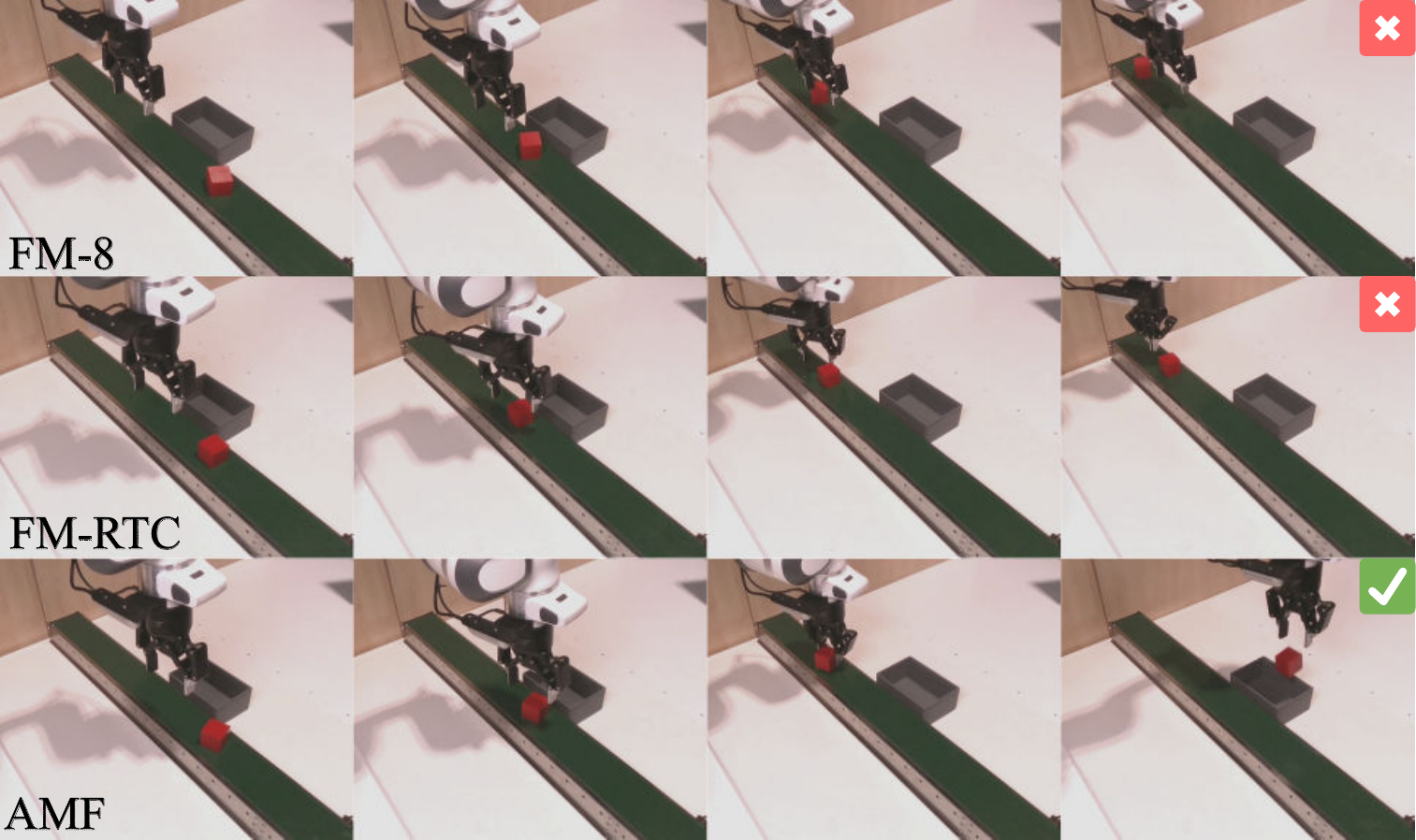}
    \caption{Rollout comparison on real-world conveyor belt task, with the maximum belt speed of 15.3 cm/s. \textbf{Top:} FM-8 initially misjudges the trajectory needed to intercept the block, and is unable to recover, eventually ending up behind the object and failing the task. \textbf{Middle:} FM with real-time chunking adjusts more aggressively, but closes the gripper prematurely and misses the object. \textbf{Bottom:} AMF tracks slightly behind initially, but performs the necessary adjustments to catch up and successfully grasp the block.}
    \label{fig:conveyor_comparison}
    \vspace{-15pt}
\end{figure}

Table~\ref{tab:conveyor_belt_results} shows success rates on the Conveyor Belt task for increasing belt speeds. The policies were trained on demonstrations with randomized speeds approximately in the range of 10--16 cm/s. While each method is quite competitive, AMF performs best overall. All the policies consistently tracked the end-effector relatively close to the block, and most failures were due to slight over- or undershoots, as exemplified in Figure~\ref{fig:conveyor_comparison}. AMF demonstrated the most consistency in executing the fine adjustments needed to fully align the grippers with the moving block. Curiously, all policies performed better at 15.3 cm/s than at 13.6 cm/s. This might indicate that some speeds were underrepresented in the data, and that this factor matters as much as the inherent increase in difficulty with higher speeds.

\begin{table}[]
    \centering
    \caption{Success rate on conveyor belt picking task. Each policy is trained for 600 epochs. We conduct experiments at 4 different belt speeds, corresponding to DC belt motor actuation with 10-13V. We perform 10 trials for each method at each speed setting, with slight variations in the initial block position on the belt.}
    \begin{tabular}[width=.48\textwidth]{lcccccc}
    \toprule
       Method   & 11.5 cm/s & 12.5 cm/s & 13.6 cm/s & 15.3 cm/s & Avg\\
       \midrule
       FM-8 & 0.70 &  \textbf{0.70} & 0.20 & 0.30 & 0.48\\
       FM-RTC & 0.50  &  0.50 & 0.10 & 0.50 & 0.40\\
       AMF  (Ours) & \textbf{0.80} & 0.50 & \textbf{0.30} & \textbf{0.60} & \textbf{0.55}\\
       \bottomrule   
    \end{tabular}
    \label{tab:conveyor_belt_results}
\end{table}

\section{Conclusion}
In this paper, we presented Adaptive Mean Flow, a robotic IL method that enables flexible, responsive, and smooth robot control. We demonstrated the efficacy of AMF on a comprehensive set of simulated and real-world tasks, both with and without dynamic components. We also presented training recipes and dedicated architectural modifications to optimize AMF performance. A limitation of the current approach is that all action modalities are treated the same. The proposed scheme was primarily designed with continuous modalities in mind, such as end-effector poses, and not discrete modes such as binary gripper commands. Indeed, a common failure mode for AMF was an inability to open/close the gripper when required. Future work will investigate modality-specific strategies to improve generalizability and robustness. There is also room to explore alternative corruption schedule formulations and parametrizations, and further improve and optimize the adaptive noise corruption scheme.

\FloatBarrier
\small
\bibliography{refs}

@article{ho2020denoising,
  title={Denoising diffusion probabilistic models},
  author={Ho, Jonathan and Jain, Ajay and Abbeel, Pieter},
  journal={Advances in neural information processing systems},
  volume={33},
  pages={6840--6851},
  year={2020}
}

@InProceedings{pmlr-v37-sohl-dickstein15,
  title = 	 {Deep Unsupervised Learning using Nonequilibrium Thermodynamics},
  author = 	 {Sohl-Dickstein, Jascha and Weiss, Eric and Maheswaranathan, Niru and Ganguli, Surya},
  booktitle = 	 {Proc. of the 32nd Int. Conf. on Machine Learning},
  pages = 	 {2256--2265},
  year = 	 {2015},
  volume = 	 {37},
}

@inproceedings{
lipman2023flow,
title={Flow Matching for Generative Modeling},
author={Yaron Lipman and Ricky T. Q. Chen and Heli Ben-Hamu and Maximilian Nickel and Matthew Le},
booktitle={The 11th Int. Conf. on Learning Representations },
year={2023},
}

@article{diffusionsurveywolf2025,
author = {Wolf, Rosa and Shi, Yitian and Liu, Sheng and Rayyes, Rania},
year = {2025},
title = {Diffusion models for robotic manipulation: a survey},
volume = {12},
journal = {Frontiers in Robotics and AI},
}

@article{chi2023diffusion,
  title={Diffusion policy: Visuomotor policy learning via action diffusion},
  author={Chi, Cheng and Xu, Zhenjia and Feng, Siyuan and Cousineau, Eric and Du, Yilun and Burchfiel, Benjamin and Tedrake, Russ and Song, Shuran},
  journal={The Int. Journal of Robotics Research},
  year={2023},
}

@INPROCEEDINGS{zhao2023learning, 
    AUTHOR    = {Tony Z. Zhao AND Vikash Kumar AND Sergey Levine AND Chelsea Finn}, 
    TITLE     = {{Learning Fine-Grained Bimanual Manipulation with Low-Cost Hardware}}, 
    BOOKTITLE = {Proc. of Robotics: Science and Systems}, 
    YEAR      = {2023},
}

@inproceedings{
liu2025bidirectional,
title={Bidirectional Decoding: Improving Action Chunking via Guided Test-Time Sampling},
author={Yuejiang Liu and Jubayer Ibn Hamid and Annie Xie and Yoonho Lee and Max Du and Chelsea Finn},
booktitle={The 13th Int. Conf. on Learning Representations},
year={2025},
}

@inproceedings{
geng2025meanflowsonestepgenerative,
title={Mean Flows for One-step Generative Modeling},
author={Zhengyang Geng and Mingyang Deng and Xingjian Bai and J Zico Kolter and Kaiming He},
booktitle={The 39th Annual Conf. on Neural Information Processing Systems},
year={2025},
}

@inproceedings{høeg2024streamingdiffusionpolicyfast,
  author={Høeg, Sigmund H. and Du, Yilun and Egeland, Olav},
  booktitle={IEEE Int. Conf. on Robotics and Automation}, 
  title={Fast Policy Synthesis with Variable Noise Diffusion Models}, 
  year={2025},
  volume={},
  number={},
  pages={4821-4828},
  doi={10.1109/ICRA55743.2025.11127858}}

@inproceedings{chenforcing2024,
 author = {Chen, Boyuan and Mart\'{\i} Mons\'{o}, Diego and Du, Yilun and Simchowitz, Max and Tedrake, Russ and Sitzmann, Vincent},
 booktitle = {Advances in Neural Information Processing Systems},
 pages = {24081--24125},
 title = {Diffusion Forcing: Next-token Prediction Meets Full-Sequence Diffusion},
 volume = {37},
 year = {2024}
}

@inproceedings{
song2021denoising,
title={Denoising Diffusion Implicit Models},
author={Jiaming Song and Chenlin Meng and Stefano Ermon},
booktitle={The 9th Int. Conf. on Learning Representations},
year={2021},
}

@inproceedings{prasadconsistency,
  author={Aaditya Prasad and Kevin Lin and Jimmy Wu and Linqi Zhou and Jeannette Bohg},
  title={Consistency Policy: Accelerated Visuomotor Policies via Consistency Distillation},
  year={2024},
  booktitle={Proc. of Robotics: Science and Systems},
}

@InProceedings{pmlr-v202-song23a,
  title = 	 {Consistency Models},
  author =       {Song, Yang and Dhariwal, Prafulla and Chen, Mark and Sutskever, Ilya},
  booktitle = 	 {Proc. of the 40th Int. Conf. on Machine Learning},
  pages = 	 {32211--32252},
  year = 	 {2023},
  volume = 	 {202},
}

@inproceedings{
wang2025onestep,
title={One-Step Diffusion Policy: Fast Visuomotor Policies via Diffusion Distillation},
author={Zhendong Wang and Max Li and Ajay Mandlekar and Zhenjia Xu and Jiaojiao Fan and Yashraj Narang and Linxi Fan and Yuke Zhu and Yogesh Balaji and Mingyuan Zhou and Ming-Yu Liu and Yu Zeng},
booktitle={42nd Int. Conf. on Machine Learning},
year={2025},
}

@inproceedings{zhangconsistencyflowpolicy2025,
author = {Zhang, Qinglun and Liu, Zhen and Fan, Haoqiang and Liu, Guanghui and Zeng, Bing and Liu, Shuaicheng},
year = {2025},
pages = {14754-14762},
title = {FlowPolicy: Enabling Fast and Robust {3D} Flow-Based Policy via Consistency Flow Matching for Robot Manipulation},
volume = {39},
booktitle = {Proc. of the AAAI Conf. on Artificial Intelligence},
}

@article{yang2024consistencyfm,
  title={Consistency Flow Matching: Defining Straight Flows with Velocity Consistency},
  author={Yang, Ling and Zhang, Zixiang and Zhang, Zhilong and Liu, Xingchao and Xu, Minkai and Zhang, Wentao and Meng, Chenlin and Ermon, Stefano and Cui, Bin},
  journal={arXiv preprint arXiv:2407.02398},
  year={2024}
}

@inproceedings{jiang2025streaming,
  title     = {Streaming Flow Policy: Simplifying diffusion/flow-matching policies by treating action trajectories as flow trajectories},
  author    = {Sunshine Jiang AND Xiaolin Fang AND Nicholas Roy AND Tom{\'a}s Lozano-P{\'e}rez AND Leslie Pack Kaelbling AND Siddharth Ancha},
  booktitle = {9th Annual Conf. on Robot Learning},
  year      = {2025},
}

@inproceedings{Shengmeanflowtames2026, title={MP1: MeanFlow Tames Policy Learning in 1-step for Robotic Manipulation}, volume={40}, booktitle={Proc. of the AAAI Conf. on Artificial Intelligence}, author={Sheng, Juyi and Wang, Ziyi and Li, Peiming and Liu, Mengyuan}, year={2026}, pages={18532-18539}}

@inproceedings{
torne2025ptp,
title={Learning Long-Context Diffusion Policies via Past-Token Prediction},
author={Marcel Torne Villasevil and Andy Tang and Yuejiang Liu and Chelsea Finn},
booktitle={9th Annual Conf. on Robot Learning},
year={2025},
}

@article{malhotra2025selfguidedactiondiffusion,
      title={Self-Guided Action Diffusion}, 
      author={Rhea Malhotra and Yuejiang Liu and Chelsea Finn},
      year={2025},
      journal={arXiv preprint arXiv:2508.12189}
}

@inproceedings{black2025realtimeexecutionactionchunking,
  title={Real-Time Execution of Action Chunking Flow Policies},
  author={Black, Kevin and Galliker, Manuel Y. and Levine, Sergey},
  booktitle={Advances in Neural Information Processing Systems},
  year={2025}
}

@article{weng2025temporalactionselectionaction,
      title={Temporal Action Selection for Action Chunking}, 
      author={Yueyang Weng and Xiaopeng Zhang and Yongjin Mu and Yingcong Zhu and Yanjie Li and Qi Liu},
      year={2025},
      journal={arXiv preprint arXiv:2511.04421}
}

@inproceedings{
liu2023flow,
title={Flow Straight and Fast: Learning to Generate and Transfer Data with Rectified Flow},
author={Xingchao Liu and Chengyue Gong and Qiang Liu},
booktitle={The 11th Int. Conf. on Learning Representations },
year={2023},
}

@inproceedings{esserscaling2024,
      title={Scaling Rectified Flow Transformers for High-Resolution Image Synthesis}, 
      author={Patrick Esser and Sumith Kulal and Andreas Blattmann and Rahim Entezari and Jonas Müller and Harry Saini and Yam Levi and Dominik Lorenz and Axel Sauer and Frederic Boesel and Dustin Podell and Tim Dockhorn and Zion English and Kyle Lacey and Alex Goodwin and Yannik Marek and Robin Rombach},
      year={2024},
      booktitle={ 41st Int. Conf. on Machine Learning},
 
}

@inproceedings{peebles2023dit,
  author={Peebles, William and Xie, Saining},
  booktitle={IEEE/CVF Int. Conf. on Computer Vision}, 
  title={Scalable Diffusion Models with Transformers}, 
  year={2023},
  volume={},
  number={},
  pages={4172-4182},
}

@inproceedings{
mclean2025metaworld,
title={Meta-World+: An Improved, Standardized, {RL} Benchmark},
author={Reginald McLean and Evangelos Chatzaroulas and Luc McCutcheon and Frank R{\"o}der and Tianhe Yu and Zhanpeng He and K.R. Zentner and Ryan Julian and J K Terry and Isaac Woungang and Nariman Farsad and Pablo Samuel Castro},
booktitle={The 39th Annual Conf. on Neural Information Processing Systems},
year={2025},
}

@inproceedings{
jia2024towards,
title={Towards Diverse Behaviors: A Benchmark for Imitation Learning with Human Demonstrations},
author={Xiaogang Jia and Denis Blessing and Xinkai Jiang and Moritz Reuss and Atalay Donat and Rudolf Lioutikov and Gerhard Neumann},
booktitle={The 12th Int. Conf. on Learning Representations},
year={2024},
}

@article{kim2026stabilizingconsistencytrainingflow,
  title={Stabilizing Consistency Training: A Flow Map Analysis and Self-Distillation},
  author={Youngjoong Kim and Duhoe Kim and Woosung Kim and Jaesik Park},
  journal={arXiv preprint arXiv:2601.22679},
  year={2026}
}

@inproceedings{he2016deep,
  title={Deep residual learning for image recognition},
  author={He, Kaiming and Zhang, Xiangyu and Ren, Shaoqing and Sun, Jian},
  booktitle={Proc. of the IEEE Conf. on computer vision and pattern recognition},
  pages={770--778},
  year={2016}
}
% \endgroup

\end{document}